%% file: main.tex
\documentclass[10pt,twocolumn,letterpaper]{article}

\usepackage[pagenumbers]{iccv} 
\usepackage{multirow}

\definecolor{iccvblue}{rgb}{0.21,0.49,0.74}
\usepackage[pagebackref,breaklinks,colorlinks,allcolors=iccvblue]{hyperref}

\def\paperID{3} 
\def\confName{ICCV}
\def\confYear{2025}

\title{SVD-ViT: Does SVD Make Vision Transformers Attend More to the Foreground?}

\author{Haruhiko Murata\\
Meijo University\\
{\tt\small harusame2357@gmail.com}
\and
Kazuhiro Hotta\\
Meijo University\\
{\tt\small 	kazuhotta@meijo-u.ac.jp}
}

\begin{document}
\maketitle
\input{sec/0_abstract}    
\input{sec/1_intro}
\input{sec/2_related}
\input{sec/3_methods}
\input{sec/4_experiments}
\input{sec/5_conclusion}
{
    \small
    \bibliographystyle{ieeenat_fullname}
    \bibliography{main}
}

\end{document}

%% file: sec/0_abstract.tex
\begin{abstract}
Vision Transformers (ViT) have been established as large-scale foundation models. However, because self-attention operates globally, they lack an explicit mechanism to distinguish foreground from background. As a result, ViT may learn unnecessary background features and artifacts, leading to degraded classification performance. To address this issue, we propose SVD-ViT, which leverages singular value decomposition (SVD) to prioritize the learning of foreground features. SVD-ViT consists of three components—\textbf{SPC module}, \textbf{SSVA}, and \textbf{ID-RSVD}—and suppresses task-irrelevant factors such as background noise and artifacts by extracting and aggregating singular vectors that capture object foreground information. Experimental results demonstrate that our method improves classification accuracy and effectively learns informative foreground representations while reducing the impact of background noise.
\end{abstract}

%% file: sec/1_intro.tex
\section{Introduction}
\label{sec:intro}

\begin{figure*}[t]
    \centering
    \includegraphics[width=0.95\linewidth]{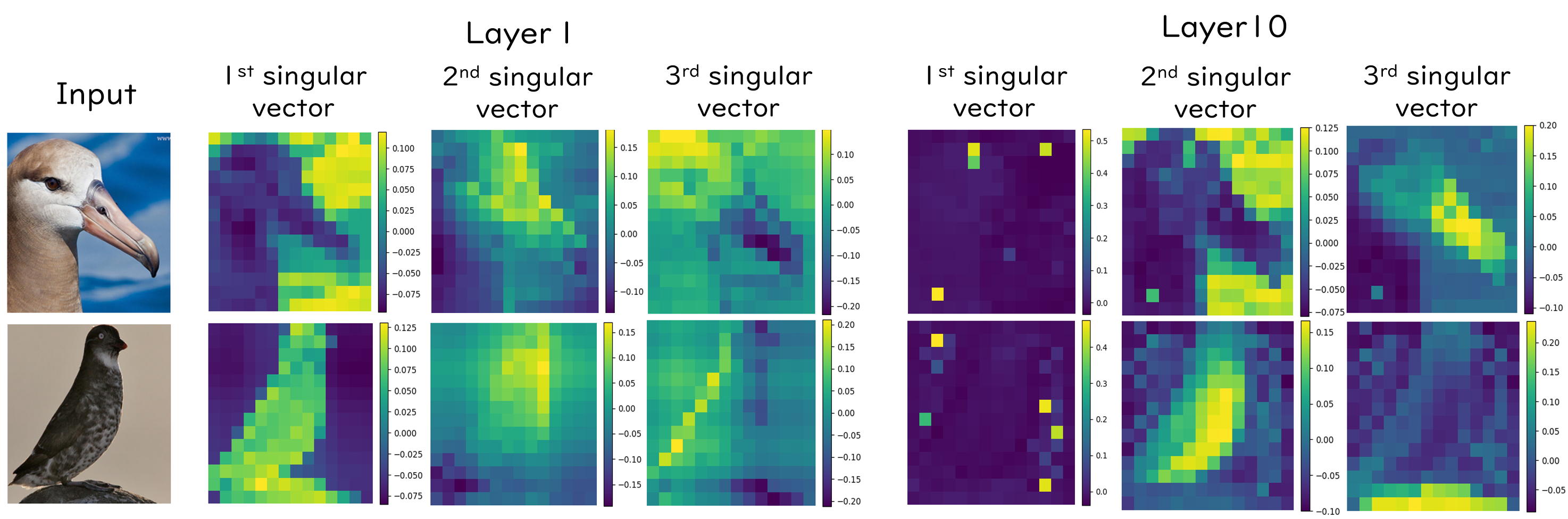}
    \caption{Visualization of the leading left singular vectors obtained by applying SVD to the patch feature matrix (number of patches $\times$ embedding dimension) at each ViT layer. Each left singular vector is reshaped to the patch grid and rendered as a heatmap. From left to right, we show the input image, Layer~1, and Layer~10; for Layer~1 and Layer~10, the first to third singular vectors (e.g., \#1--\#3) are displayed. In shallow layers, components corresponding to the foreground emerge, whereas in deeper layers, spike-like high-norm components are observed.}
    \label{fig:svd_vis}
\end{figure*}

In recent years, Vision Transformers (ViT)~\cite{vit} have been established as standard backbone models in computer vision. ViT performs classification by feeding a learnable $[\mathrm{CLS}]$ token into a linear layer. Existing visualization methods~\cite{attentionattribute} have reported that the $[\mathrm{CLS}]$ token exhibits a high correlation with foreground regions. However, because the self-attention mechanism—the core of ViT—is based on similarity computation among patches, it does not explicitly incorporate structural inductive biases that encourage foreground/background separation, such as the locality commonly observed in CNN. As a result, depending on the task, background noise may contaminate feature representations and lead to degraded classification accuracy.

To address this issue, we propose an approach that introduces singular value decomposition (SVD) into the feature extraction process of ViT, based on the hypothesis that low-rank approximation via SVD contributes to separating foreground and background. SVD has the property of concentrating a matrix’s variance into a small number of leading singular values. In the context of image processing, if we assume that “the foreground exhibits large variance in the feature space, whereas the background exhibits relatively small variance,” then SVD can be used to separate the two.

Before designing our method, we applied SVD to intermediate feature maps of a model obtained by fine-tuning a pretrained ViT on a downstream task (i.e., transfer learning), and visualized the leading singular vectors. As shown in Fig.~\ref{fig:svd_vis}, we obtained the following two observations:
\begin{itemize}[nosep,leftmargin=*]
  \item The leading singular vectors of ViT feature maps capture foreground regions, and we also confirmed vectors that locally correspond to fine-grained parts of the foreground.
  \item In deeper layers, we observed high-norm artifacts, as pointed out by Darcet et al.~\cite{registervit}.
\end{itemize}

Based on these observations, we propose the \textbf{SPC module}, which generates a $[\mathrm{SPC}]$ token as an alternative to the conventional $[\mathrm{CLS}]$ token by leveraging singular vectors that capture foreground regions. By utilizing leading singular vectors in which foreground information is concentrated, the SPC module acquires robust feature representations while suppressing background noise. We refer to the ViT equipped with the SPC module as \textbf{SVD-ViT}. Furthermore, we propose two plugin modules, \textbf{SSVA} and \textbf{ID-RSVD}, to extend and strengthen the functionality of the SPC module.

The novelty of our method lies in directly integrating SVD into the feature extraction process. Traditionally, applications of SVD in deep learning have mainly focused on adjusting static weight parameters, such as model compression~\cite{denton} and parameter-efficient fine-tuning (PEFT)~\cite{salt}. In contrast, our work differs from prior studies in that it applies SVD to feature maps, thereby suggesting new possibilities for spectral analysis in feature space.

We evaluate the proposed SVD-ViT on five major image recognition benchmark datasets (CUB-200-2011, FGVC-Aircraft, Stanford Cars, Food-101, and CIFAR-100) to verify its effectiveness. Experimental results show that our method achieves higher classification accuracy than the ViT baseline across all datasets. In particular, we observe notable improvements of up to +2.38 percentage points on CUB-200-2011 and +2.82 percentage points on FGVC-Aircraft.

Our contributions are summarized as follows:
\begin{itemize}
    \item We perform SVD and visualization on feature maps of pretrained ViT and demonstrate that leading singular components strongly correlate with foreground regions.
    \item We propose a new architecture (SVD-ViT) that incorporates SVD within ViT layers to emphasize foreground features and improve robustness to background noise.
    \item We introduce a new perspective of applying SVD to the feature extraction process, highlighting its novelty for analyzing ViT feature representations.
\end{itemize}

%% file: sec/2_related.tex
\section{Related Work}

\subsection{Limitations of Vision Transformer}
Vision Transformer (ViT)~\cite{vit} treats an image as a sequence of patches and achieves strong performance by learning global relationships via self-attention.
For classification, a learnable $[\mathrm{CLS}]$ token aggregates information from patch tokens and is used for the final prediction.
However, since self-attention is based on similarity among patches, it lacks an inductive bias that explicitly separates foreground and background, and information originating from the background can be aggregated.
Furthermore, recent studies have reported that input-independent high-norm tokens (artifacts) can emerge in the feature space of ViT and may affect attention distributions and feature aggregation~\cite{registervit,register_small,register_free,register_mamba}.
To suppress these undesirable components, we introduce the SPC module (\S\ref{subsec:svd-vit}), which generates an aggregation token using spatial singular vectors obtained by spectral decomposition of feature maps, thereby extending feature aggregation in ViT.
In addition, we improve the selectivity of the subspace used for aggregation via a plugin module (SSVA; \S\ref{subsec:ssva}) that integrates singular vectors through a weighted sum.

Our novelty lies in \textbf{using SVD for feature extraction}.
Representative applications of SVD in deep learning include model compression via low-rank approximation of weights~\cite{denton} and parameter-efficient fine-tuning (PEFT) using SVD-structured parameterization~\cite{adalora,soma,salt}, which primarily target static weight parameters.
In contrast, our work differs from prior studies in that it performs token generation and integration using SVD on dynamic intermediate features.

\subsection{RSVD with Fixed Sketching and Its Limitations}
\begin{figure}[t]
    \centering
    \includegraphics[width=0.95\linewidth]{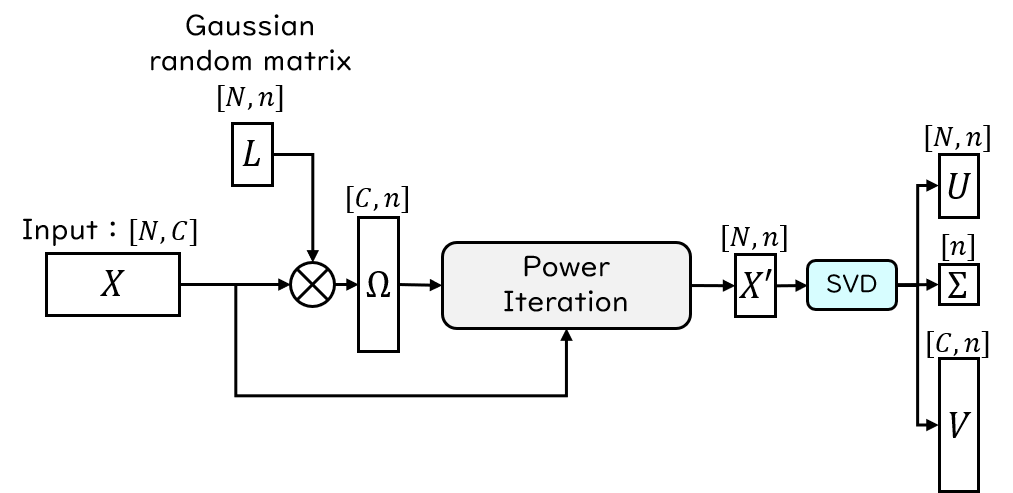}
    \caption{Overview of RSVD. A low-rank approximation matrix is constructed via randomized sketching and iterative orthogonalization, and applying SVD to the resulting matrix enables extracting only the leading singular vectors.}
    \label{fig:rsvd}
\end{figure}

Standard SVD is computationally expensive for large matrices, making direct application to feature maps in deep learning burdensome.
To address this issue, Halko et al.~\cite{halko2011} proposed \textbf{Randomized SVD (RSVD)} based on random projections and systematized a framework for efficiently approximating leading singular components.
Hereafter, we refer to the method of Halko et al. as RSVD.

RSVD projects an input matrix $X$ to a low-dimensional space using a random matrix $\Omega$ and constructs a smaller matrix $Y = X\Omega$ (\textbf{sketching}).
Because random projections preserve the dominant subspace with high probability, RSVD can efficiently approximate the leading subspace of $X$.
It then optionally improves the approximation via power iteration with iterative orthogonalization, producing a low-rank approximation matrix $X'$; finally, the leading singular components are obtained by performing SVD on $X'$ (Fig.~\ref{fig:rsvd}).

In the context of image recognition, however, leading singular components (i.e., directions with large variance) are not necessarily discriminative.
For example, when the background region exhibits large variance, or when RSVD is applied in deeper layers where artifacts are more likely to appear, background noise or artifacts may emerge in the leading components (Fig.~\ref{fig:svd_vis}).
Moreover, in fine-grained image classification, it is plausible that localized part features appearing in lower components can be more important than foreground components appearing in the leading components.
Therefore, simply applying RSVD may not stably extract singular directions required for the task.

To address this, we propose a method (ID-RSVD; \S\ref{subsec:id-rsvd}) that selectively emphasizes task-useful subspaces by replacing the fixed random sketch matrix with a learnable matrix generated in an input-dependent manner.

%% file: sec/3_methods.tex
\section{Proposed Method}
In this section, we propose a new architecture, \textbf{SVD-ViT}, which aims to improve recognition accuracy by encouraging foreground--background separation in image recognition.
While ViT excels at extracting global representations, it also tends to incorporate background-derived noise.
To address this issue, we integrate singular value decomposition (Singular Value Decomposition; SVD) into the model to emphasize foreground information.
Below, we first describe the motivation based on an analysis of the feature space, and then detail the core component of our method, the \textbf{SPC module}, as well as the plugin modules \textbf{SSVA} and \textbf{ID-RSVD}.

\subsection{Analysis of ViT Feature Maps}
\label{subsec:svd_vis}
To examine the hypothesis that foreground information is dominant in intermediate features of ViT, we conduct spectral analysis using RSVD.
Let $N$ denote the number of patch tokens and $C$ the embedding dimension.
We represent the intermediate features including the CLS token as $\mathbf{X}\in\mathbb{R}^{(1+N)\times C}$.
By approximately extracting the top $n$ singular components via RSVD, we obtain the following low-rank approximation:
\begin{equation}
  \mathbf{X} \approx \mathbf{U}_n \mathbf{\Sigma}_n \mathbf{V}_n^\top .
  \label{eq:rsvd}
\end{equation}
Here, $\mathbf{\Sigma}_n\in\mathbb{R}^{n\times n}$ is the diagonal matrix of singular values and $\mathbf{V}_n\in\mathbb{R}^{C\times n}$ is the matrix of right singular vectors.
The left singular vectors $\mathbf{U}_n$ capture dominant directions over tokens (spatial locations).
For visualization, we discord the row of $\mathbf{U}_n$ corresponding to $[\mathrm{CLS}]$ and reshape each column vector into a 2D grid according to the original patch layout, which allows us to inspect activated regions.

Figure~\ref{fig:svd_vis} shows visualization results of the leading singular vectors at a shallow layer (Layer~1) and a deep layer (Layer~10) using CUB-200-2011~\cite{cub}.
From these results, we obtain the following observations.

First, we observe the \textbf{dominance of foreground information}.
Regardless of layer depth, the leading singular vectors tend to capture the main subject (foreground).
This supports the hypothesis that ``semantically important information exhibits large variance in the feature space.''

Second, we observe \textbf{contamination by background noise and artifacts}.
Because SVD extracts directions with large variance, background information can appear as leading components when the background is complex.
Moreover, we observe high-norm artifacts that do not carry semantic information, as pointed out by Darcet et al.~\cite{registervit}.
Therefore, rather than naively using the leading components, it is important to incorporate a mechanism (SSVA, described later) that selectively emphasizes components that are useful for the task.

\subsection{SVD-ViT Architecture and Details of the SPC Module}
\label{subsec:svd-vit}
\begin{figure}[t]
    \centering
    \includegraphics[width=0.95\linewidth]{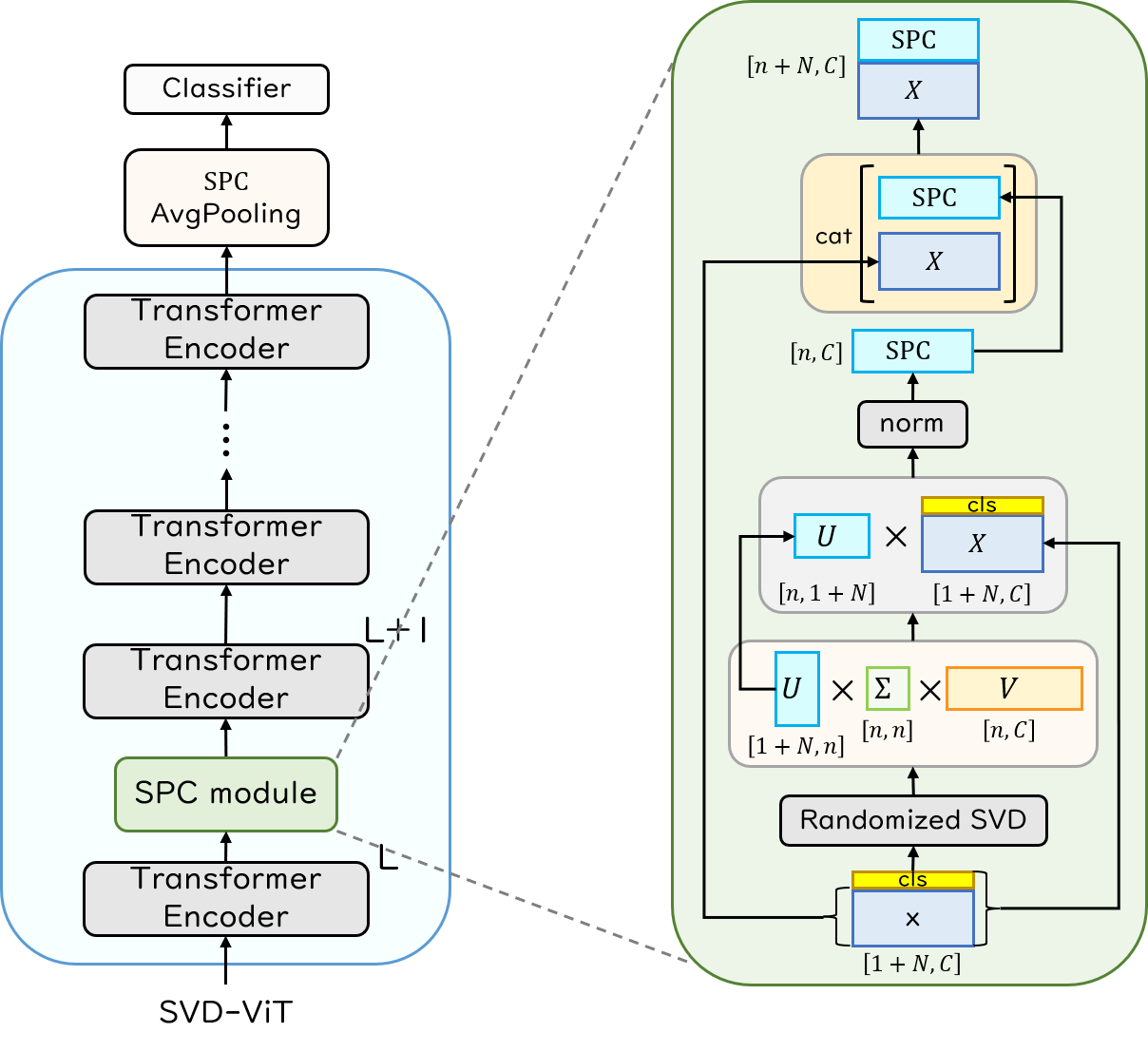}
    \caption{Overview of SVD-ViT.}
    \label{fig:svdvit}
\end{figure}

As shown in \S\ref{subsec:svd_vis}, the leading singular components of intermediate features tend to capture foreground regions.
Based on this observation, we regard the leading spatial singular vectors as spatial principal components and propose the \textbf{SPC module}, which generates an aggregation token, the $[\mathrm{SPC}]$ token, that is useful for classification.
We refer to the ViT model equipped with the SPC module as \textbf{SVD-ViT}; an overview is shown in Fig.~\ref{fig:svdvit}.

The SPC module functions as a plugin inserted between Transformer encoder blocks.
Let the input token sequence including $[\mathrm{CLS}]$ be $\mathbf{X}\in\mathbb{R}^{(1+N)\times C}$, and write $\mathbf{X}=[\mathbf{x}_{cls};\,\mathbf{X}_p]$.
Here, $\mathbf{x}_{cls}\in\mathbb{R}^{1\times C}$ is the $[\mathrm{CLS}]$ token and $\mathbf{X}_p\in\mathbb{R}^{N\times C}$ is the patch token sequence.
We apply RSVD to $\mathbf{X}$ and obtain the top $n$ left singular vectors $\mathbf{U}_n\in\mathbb{R}^{(1+N)\times n}$.
We then generate $n$ $[\mathrm{SPC}]$ tokens, which are likely to contain foreground information, by projecting $\mathbf{X}$ onto the subspace spanned by $\mathbf{U}_n$:
\begin{equation}
  \mathbf{S} = \mathbf{U}_n^{\top}\mathbf{X} \in \mathbb{R}^{n\times C},
  \qquad
  \mathbf{SPC} = \mathrm{LN}(\mathbf{S}) .
  \label{eq:spc}
\end{equation}
Here, $\mathrm{LN}(\cdot)$ denotes Layer Normalization.
We append the resulting $\mathbf{SPC}\in\mathbb{R}^{n\times C}$ to the patch token sequence, and feed the next Transformer encoder with
$\tilde{\mathbf{X}}=[\mathbf{SPC};\,\mathbf{X}_p]\in\mathbb{R}^{(n+N)\times C}$, i.e., the token sequence without $[\mathrm{CLS}]$.
This enables subsequent layers to perform attention while referencing the foreground-enhanced $[\mathrm{SPC}]$ tokens.
Finally, we average-pool the $\mathbf{SPC}\in\mathbb{R}^{n\times C}$ obtained at the last layer along the token dimension ($n$) and feed it into the classifier to obtain class predictions.

\subsection{SSVA: Spatial Singular Value Aggregator}
\label{subsec:ssva}
\begin{figure}[t]
    \centering
    \includegraphics[width=0.95\linewidth]{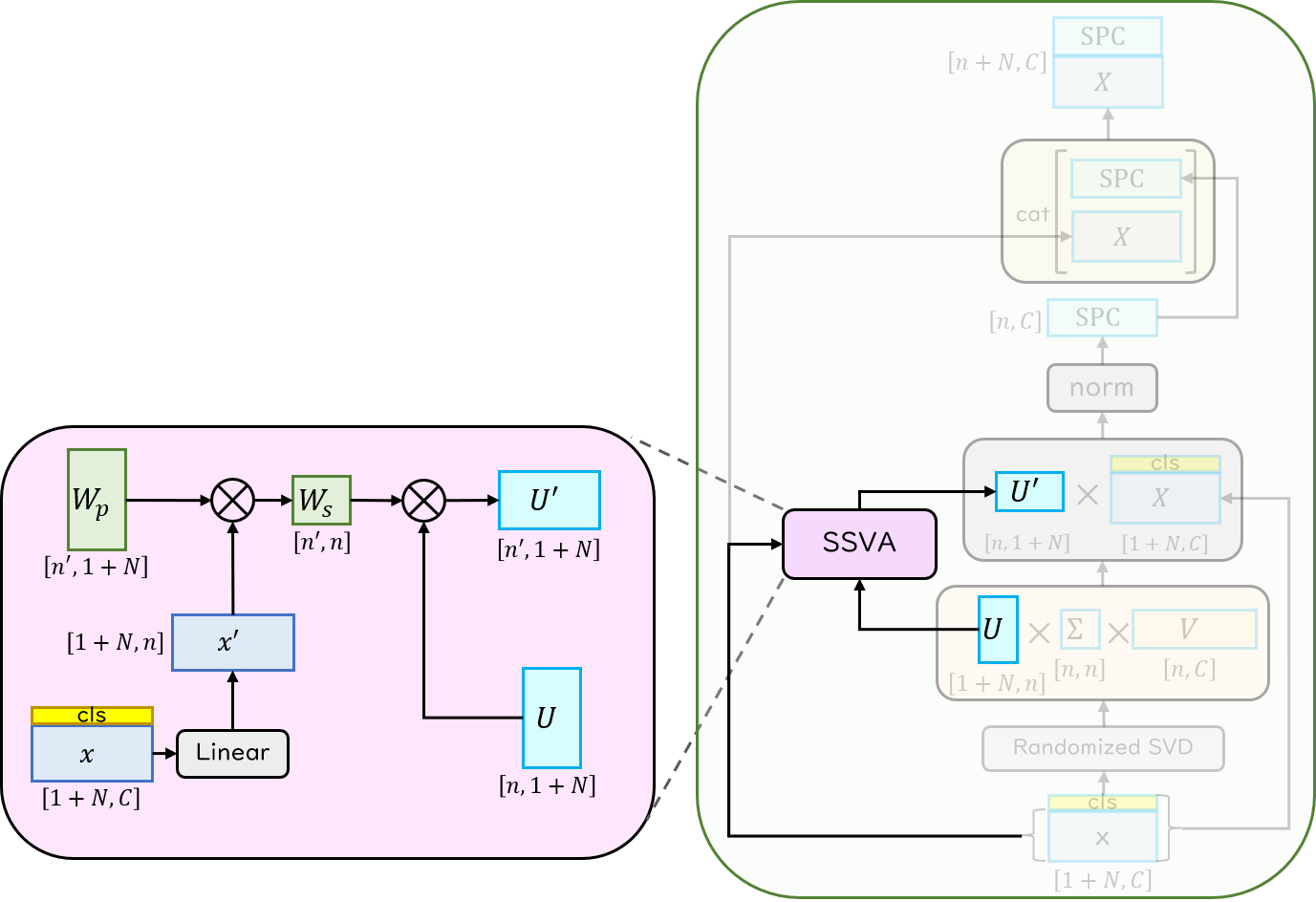}
    \caption{Overview of SSVA.}
    \label{fig:ssva}
\end{figure}

In the SPC module, the top $n$ singular vectors obtained by SVD are used as $[\mathrm{SPC}]$ tokens.
However, components with large variance are not always useful for classification.
Therefore, we introduce \textbf{Spatial Singular Value Aggregator (SSVA)}, which selects important components from the extracted $n$ singular vectors in an input-dependent manner and aggregates them into $n'$ bases (Fig.~\ref{fig:ssva}).

\paragraph{Formulation.}
Let $\mathbf{X}\in\mathbb{R}^{(1+N)\times C}$ denote the token features of ViT including CLS.
We obtain the left singular vectors $\mathbf{U}_n\in\mathbb{R}^{(1+N)\times n}$ from $\mathbf{X}$ via randomized SVD.
In SSVA, we use $\bar{\mathbf{U}}_n=\mathbf{U}_n^\top\in\mathbb{R}^{n\times(1+N)}$ to facilitate component mixing.

First, we linearly project $\mathbf{X}$ to generate a query $\mathbf{X}'\in\mathbb{R}^{(1+N)\times n}$:
\begin{equation}
  \mathbf{X}' = \mathrm{Linear}(\mathbf{X}) .
\end{equation}
Next, using learnable parameters $\mathbf{W}_p\in\mathbb{R}^{n'\times(1+N)}$, we compute weights
$\mathbf{W}_s\in\mathbb{R}^{n'\times n}$ that determine which singular vectors each aggregated token should attend to:
\begin{equation}
  \mathbf{W}_s = \mathbf{W}_p \mathbf{X}' .
  \label{eq:ssva1}
\end{equation}
$\mathbf{W}_s$ is a selection weight that represents ``which singular vectors to mix and to what extent'' based on the input features.
Using these weights, we aggregate the singular vectors to obtain reconstructed bases $\mathbf{U}'_n\in\mathbb{R}^{n'\times(1+N)}$:
\begin{equation}
  \mathbf{U}'_n = \mathbf{W}_s \bar{\mathbf{U}}_n .
  \label{eq:ssva2}
\end{equation}
Finally, we project $\mathbf{X}$ with $\mathbf{U}'_n$ to generate $n'$ $[\mathrm{SPC}]$ tokens
$\mathbf{Z}_{\mathrm{spc}}\in\mathbb{R}^{n'\times C}$:
\begin{equation}
  \mathbf{Z}_{\mathrm{spc}} = \mathbf{U}'_n \mathbf{X} .
  \label{eq:ssva3}
\end{equation}
Equations~\eqref{eq:ssva1}--\eqref{eq:ssva3} show that SSVA dynamically integrates important singular directions according to the input and constructs task-useful $[\mathrm{SPC}]$ tokens.

\begin{figure}[t]
    \centering
    \includegraphics[width=0.95\linewidth]{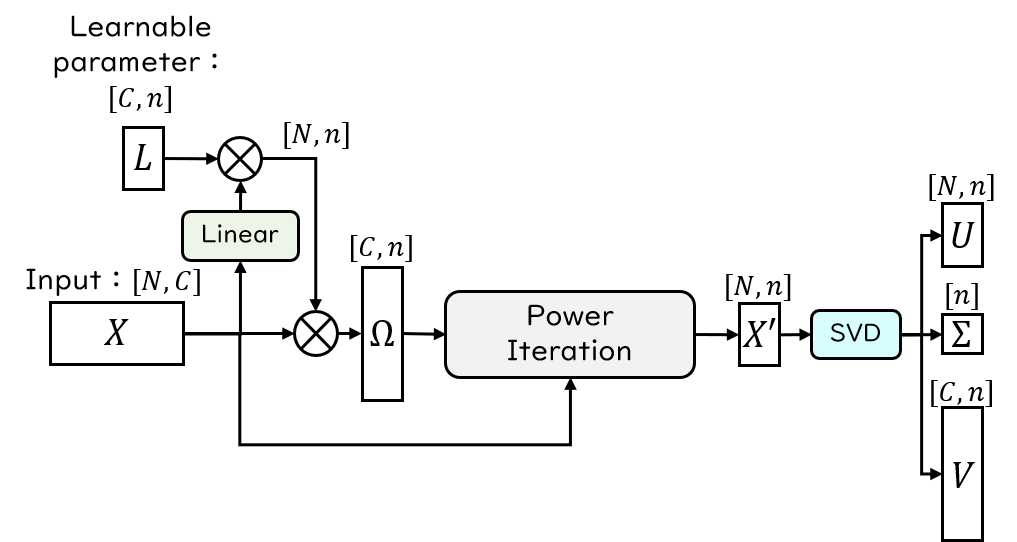}
    \caption{Overview of ID-RSVD.}
    \label{fig:id-rsvd}
\end{figure}

\subsection{ID-RSVD: Input-Dependent Randomized SVD}
\label{subsec:id-rsvd}
In RSVD~\cite{halko2011}, the projection matrix $\mathbf{\Omega}$ is used as an input-independent mapping from input features $\mathbf{X}$ to a low-dimensional subspace.
Following a common setup, we treat $\mathbf{\Omega}$ as a Gaussian random matrix.
While this projection theoretically yields a good subspace with high probability, in image recognition the leading singular components are not necessarily discriminative.
For example, when the background region exhibits large variance, or when RSVD is applied in deep layers where artifacts are likely to appear, background noise or artifacts may contaminate the leading components (Fig.~\ref{fig:svd_vis}).
Moreover, in fine-grained image classification, localized part features appearing in lower components can be more important than global foreground information appearing in the leading components.
Therefore, simply applying RSVD may not stably extract the singular directions required for the task.

To address this issue, we propose \textbf{Input-Dependent RSVD (ID-RSVD)}, which extends the random sketching process to be input-dependent and learnable.
As illustrated in Fig.~\ref{fig:id-rsvd}, ID-RSVD dynamically generates the projection matrix $\mathbf{\Omega}$ conditioned on the input $\mathbf{X}$, and further guides the search directions toward task-useful subspaces via learnable parameters, aiming to estimate subspaces that are less dominated by background noise.
\begin{equation}
  \mathbf{\Omega} = \phi(\mathbf{X})\mathbf{L}
  \label{eq:id_rsvd1}
\end{equation}
Here, $\phi(\cdot)$ denotes a linear transformation (a linear layer), and $\mathbf{L}$ is a learnable parameter that determines the search directions.
By generating $\mathbf{\Omega}$ from the input features, we obtain a sketch matrix tailored to each input; by learning $\mathbf{L}$, we can preferentially search for discriminative singular directions regardless of the magnitude of variance.

We then improve the approximation accuracy via iterative computation using the power iteration method.
By combining this iterative refinement with dynamic generation of the projection matrix, ID-RSVD can stably estimate task-useful subspaces even under limited rank settings, enabling accurate feature extraction.

%% file: sec/4_experiments.tex
\begin{table}[t]
    \centering
    \setlength{\tabcolsep}{4pt}
    \caption{Layer-wise search results for the insertion position on CUB-200-2011. Diff (pt) indicates the accuracy difference from ViT (CLS=1), rounded to two decimal places.}
    \label{tab:result_cub}
    \begin{tabular}{lccc}
        \toprule
        Model & Layer & Accuracy (\%) & Diff (pt) \\
        \midrule
        ViT (CLS=1) & -- & 84.50 & -- \\
        ViT (CLS=8) & -- & 84.64 & $+0.14$ \\
        \midrule
        \multirow{13}{*}{SVD-ViT (SPC)}
            & 0  & \textcolor{blue}{84.09} & $-0.41$ \\
            & 1  & 84.54 & $+0.03$ \\
            & 2  & 85.14 & $+0.64$ \\
            & 3  & 84.73 & $+0.22$ \\
            & 4  & 84.97 & $+0.47$ \\
            & 5  & 84.98 & $+0.48$ \\
            & 6  & 85.05 & $+0.55$ \\
            & 7  & 85.35 & $+0.85$ \\
            & 8  & 85.92 & $+1.42$ \\
            & 9  & 85.85 & $+1.35$ \\
            & 10 & 85.61 & $+1.10$ \\
            & \textbf{11} & \textcolor{red}{\textbf{86.74}} & \textbf{$+2.24$} \\
            & \textcolor{gray}{12} & \textcolor{gray}{42.18} & \textcolor{gray}{$-42.32$} \\
        \bottomrule
    \end{tabular}
\end{table}

\section{Experiments}
\subsection{Experimental Setup}

\begin{table*}[!t]
    \centering
    \caption{Comparison of classification accuracy between the proposed method and the baseline on five benchmark datasets.
The proposed method (SVD-ViT) achieves the best accuracy on all datasets and outperforms the baseline.
In particular, it improves accuracy by up to 2.52 points on CUB200 and 2.82 points on FGVC-Aircraft, suggesting that suppressing artifacts and noise in the feature space via SVD can contribute to better recognition performance.}
    \label{tab:result_comparison}
    \renewcommand{\arraystretch}{1.2}
    \setlength{\tabcolsep}{4pt}
    \begin{tabular}{llcccccc}
        \toprule
        \multicolumn{2}{c}{Method} & Layer & CUB-200 & FGVC & Stanford & Food & CIFAR \\
        \multicolumn{2}{c}{}       &       & -2011   & Aircraft & Cars & 101 & 100 \\
        \midrule
        \multirow{2}{*}{ViT} & CLS=1 & - & 84.50 & 77.86 & 89.03 & 89.68 & 86.88 \\
                             & CLS=8 & - & 84.64 & 76.27 & 88.60 & 89.54 & 86.79 \\
        \midrule
        \multirow{12}{*}{SVD-ViT} & \multirow{3}{*}{SPC} 
                                   & 9 & 85.85 & 78.61 & 89.80 & 89.86 & 87.32 \\
                                  && 10 & 85.61 & 79.27 & 88.31 & 89.64 & 87.42 \\
                                  && 11 & 86.74 & \textcolor{red}{\textbf{80.68}} & 89.79 & 89.66 & 87.09 \\
        \cmidrule{2-8}
                                  & \multirow{3}{*}{SPC + SSVA} 
                                   & 9 & 86.47 & 79.45 & 89.65 & 89.86 & \textcolor{red}{\textbf{88.05}} \\
                                  && 10 & 86.54 & 79.84 & 89.40 & 89.68 & 87.72 \\
                                  && 11 & 86.37 & 79.63 & 89.70 & \textcolor{red}{\textbf{90.07}} & 87.80 \\
        \cmidrule{2-8}
                                  & \multirow{3}{*}{\shortstack[l]{SPC + \\ ID-RSVD}} 
                                   & 9 & 85.40 & 79.45 & 89.03 & 89.95 & 87.16 \\
                                  && 10 & 86.33 & 78.46 & 88.22 & 89.75 & 86.83 \\
                                  && 11 & 86.45 & 79.93 & 89.43 & 89.56 & 87.40 \\
        \cmidrule{2-8}
                                  & \multirow{3}{*}{\shortstack[l]{SPC + SSVA \\ + ID-RSVD}} 
                                   & 9 & 86.18 & 77.74 & 89.58 & 90.02 & 87.29 \\
                                  && 10 & \textcolor{red}{\textbf{87.02}} & 78.82 & \textcolor{red}{\textbf{89.85}} & 89.75 & 87.60 \\
                                  && 11 & 86.30 & 78.10 & 89.77 & 90.02 & 87.66 \\
        \bottomrule
    \end{tabular}
\end{table*}

We evaluate our method on CIFAR-100, which is widely used for general image classification, and on four fine-grained image classification benchmarks: Food-101 \cite{food101}, FGVC-Aircraft \cite{aircraft}, CUB-200-2011 \cite{cub}, and Stanford Cars \cite{stanford_cars}.
As shown in Fig.~\ref{fig:svd_vis}, singular vectors of the features can capture different spatial structures within the foreground. Therefore, we expect the proposed SPC module to help recognize detailed spatial information.
For this reason, we perform experiments not only on a general dataset but also on fine-grained image classification datasets.

We use full fine-tuning in all experiments.
Because our method inserts the SPC module into intermediate layers of a pretrained ViT and changes the token structure and feature space, updating only the final layer may not adapt well to these changes.
Thus, we update the entire model to encourage adaptation to the new feature aggregation scheme.

As the baseline, we use ViT-Base/16 pretrained on ImageNet-21k \cite{imagenet}, implemented in timm \cite{timm}.
For all experiments, we train for 50 epochs with a batch size of 128 and use Adam as the optimizer.
We use cosine annealing for the learning rate schedule: after a 5-epoch warm-up, the learning rate increases to a maximum of $1\times10^{-4}$ and then decays.
We use cross-entropy loss.
All experiments are conducted on an NVIDIA Quadro RTX 8000 GPU.
For the SPC module, we set the number of singular vectors extracted by RSVD to $n=8$ and the number of aggregated tokens after SSVA to $n'=4$.

\begin{figure*}[!th]
    \centering
    \includegraphics[width=0.95\linewidth]{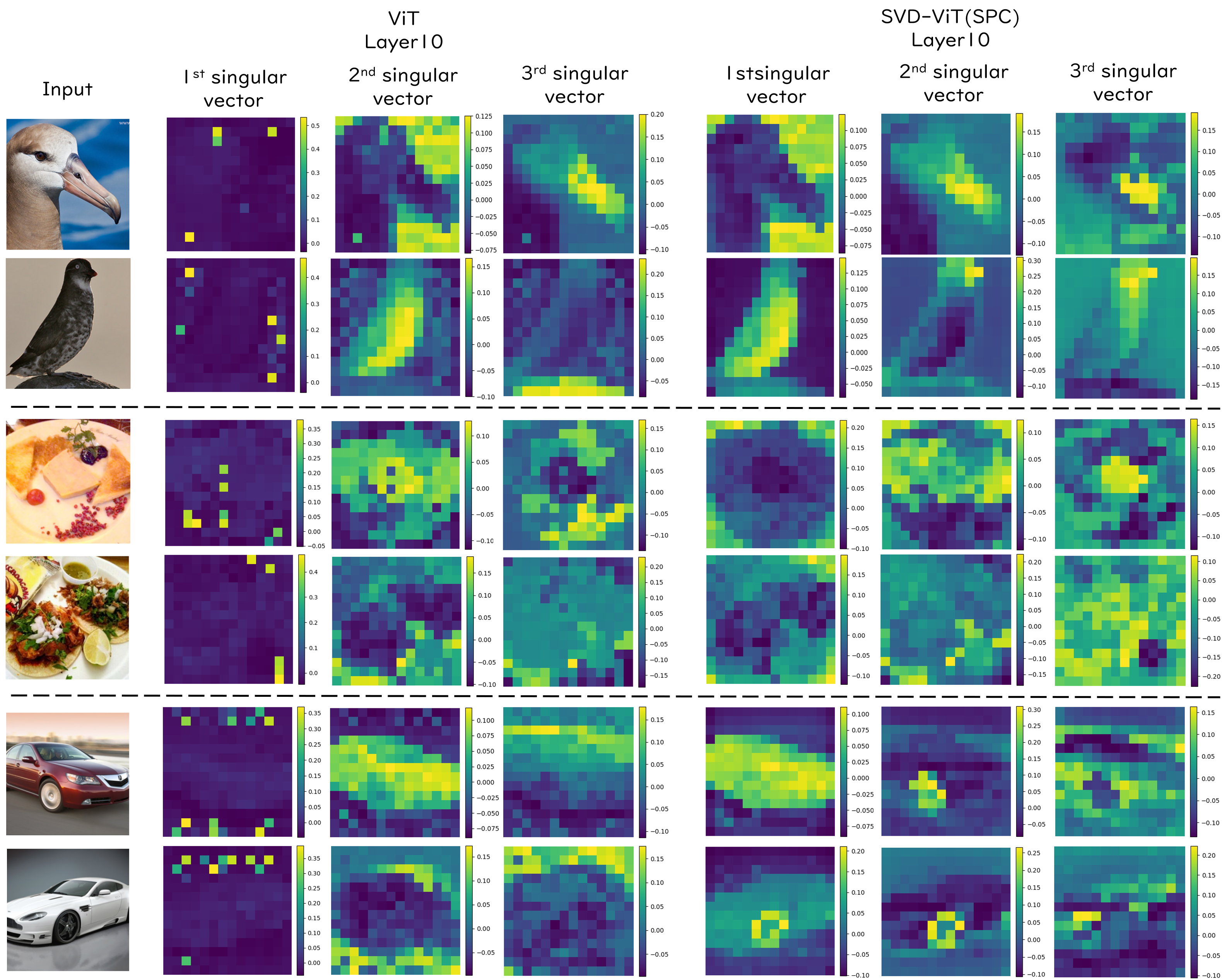}
    \caption{Comparison of top singular-vector visualizations for intermediate features of ViT and SVD-ViT on CUB-200-2011.
In ViT, some components show spike-like high-norm patches (artifacts).
In contrast, in SVD-ViT with only the SPC module (without additional plugins such as SSVA/ID-RSVD), the top vectors show spatial structures along the foreground and local object parts, and the artifacts are suppressed.}
    \label{fig:spc_vis}
\end{figure*}

\subsection{Results}
Tables~\ref{tab:result_cub} and~\ref{tab:result_comparison} summarize the results.
On CUB-200-2011, we search the insertion layer for the SPC module. Inserting it after Layer 11 is the most effective and achieves 86.74\% accuracy, which is an improvement of \textbf{+2.24\,pt} over the baseline (ViT, CLS=1) (Table~\ref{tab:result_cub}).
In contrast, inserting it right after the final layer (Layer 12) causes a large drop in accuracy, suggesting that replacing the token structure just before the classification head can destabilize training.
Overall, we observe larger gains in deeper layers than in shallower layers, indicating that SPC-based foreground enhancement is more effective on semantically abstract feature spaces.

Next, based on the above observation, we compare multiple datasets using Layers 9--11 as candidates (Table~\ref{tab:result_comparison}).
SVD-ViT achieves the best accuracy and outperforms the baseline on all datasets.
In particular, on CUB200, SPC+SSVA+ID-RSVD at Layer 10 achieves 87.02\% (+2.52\,pt over the baseline), and on FGVC-Aircraft, SPC alone at Layer 11 achieves 80.68\% (+2.82\,pt).
These results suggest that extracting singular components based on SVD can help capture foreground information that is useful for recognition, especially when background noise and artifacts are strong.

Across the five datasets (Table~\ref{tab:result_comparison}), SVD-ViT achieves accuracy that is higher than or comparable to the baseline (ViT, CLS=1) on all datasets.
In particular, on CUB200, SPC+SSVA+ID-RSVD reaches 87.02\%, showing the effectiveness of our method for fine-grained image classification.
On FGVC-Aircraft, SPC alone performs best (80.68\%, +2.82\,pt), suggesting that the additional modules are not always consistently beneficial.

\subsection{Qualitative Evaluation}
From the qualitative results in Fig.~\ref{fig:spc_vis}, we observe that the top singular vectors of ViT sometimes contain local spikes that do not correspond to the object shape.
This is consistent with prior reports that input-irrelevant high-norm tokens (artifacts) can appear in the feature space~\cite{registervit,register_small,register_free,register_mamba}, which may destabilize feature aggregation and attention distributions.
In contrast, in SVD-ViT with the SPC module, the top components tend to appear as continuous structures aligned with the foreground, and spike-like responses are relatively suppressed.
This trend is consistent with the idea that SPC re-parameterizes features by projecting them onto a leading subspace, thereby emphasizing foreground-related variance structures rather than outlier-like local components.
Similar trends were observed across multiple datasets, suggesting that SPC-based foreground emphasis may work in a dataset-agnostic manner.
However, we also observed that the sign of the singular vectors can flip across inputs in the visualizations, leading to inconsistent positive/negative polarity.

\subsection{Analysis}
\textbf{Sign ambiguity of SVD.}
For the singular value decomposition $\mathbf{X}=\mathbf{U}\mathbf{\Sigma}\mathbf{V}^\top$, for any diagonal matrix $\mathbf{D}$ whose diagonal entries are $\pm 1$, the following identity holds:
\begin{equation}
\mathbf{X} = (\mathbf{U}\mathbf{D})\mathbf{\Sigma}(\mathbf{V}\mathbf{D})^\top .
\end{equation}
That is, flipping the signs of the left and right singular vectors simultaneously for each singular component does not change the decomposition, so the signs of singular vectors are not uniquely determined.

Due to this property, in the visualization in Fig.~\ref{fig:spc_vis}, whether the foreground appears as positive or negative can vary across inputs.
Moreover, in numerical computation, the sign choice can depend on the execution environment and implementation because of randomness in RSVD and rounding errors from parallel computation.
Since our method generates tokens based on singular vectors and feeds them to subsequent attention layers, such sign inconsistencies can reduce the consistency of the feature representation and may cause unstable training.
Therefore, for stable use in deep learning, it is desirable to introduce constraints or mechanisms that enforce consistent signs of singular vectors across inputs.

\textbf{Effect of the insertion layer.}
In the insertion-layer search on CUB-200-2011 (Table~\ref{tab:result_cub}), inserting SPC after Layer 11 was most effective, while inserting it right after the final layer caused a large accuracy drop.
Deeper layers tend to form more semantically abstract representations, where variance related to the foreground/category becomes more coherent.
Thus, applying SPC projection at this stage may relatively reduce the impact of background-induced variation and promote foreground-based aggregation.
In contrast, inserting the SPC module after the final layer may break the compatibility with the representation distribution assumed by the classifier head, causing a distribution shift and potential instability.

\textbf{Varying effectiveness of additional modules.}
As shown in Table~\ref{tab:result_comparison}, the gains from SSVA and ID-RSVD vary across datasets, and they are not always effective.
One possible reason is that both modules perform input-dependent weighting (mixing/projection based on inner products); if the input features dominate too strongly, the intended behavior---selective emphasis of singular directions (SSVA) and searching sketch directions suitable for the input (ID-RSVD)---may not work sufficiently.
In such cases, the selection of important components can become unstable, and the model may again become more susceptible to background and noise.

\subsection{Limitations and Future Work}
Our evaluation is based on full fine-tuning with a pretrained model as initialization.
To more strongly claim that SVD captures components useful for the foreground, it will be necessary to validate the method under training from scratch.
In addition, the extra modules (SSVA and ID-RSVD) show dataset-dependent effects and do not always guarantee improvements (Table~\ref{tab:result_comparison}).
There is room to improve their design and computation, and future work will focus particularly on better ways to introduce input dependence for more stable performance gains.
Furthermore, the SPC module, SSVA, and ID-RSVD currently do not use the right singular vectors or singular values.
In future work, we will investigate approaches that leverage principal directions in the feature space, as well as selection/weighting of singular vectors based on singular values, aiming to balance the expressiveness and stability of SVD-based feature aggregation.

%% file: sec/5_conclusion.tex
\section{Conclusion}
In this study, we proposed \textbf{SVD-ViT}, an architecture that integrates singular value decomposition (SVD) into the feature extraction process of Vision Transformers (ViT) to emphasize and aggregate foreground information.
The core component, the \textbf{SPC module}, generates aggregation tokens based on a leading subspace of intermediate features, providing feature representations that suppress background-induced variation and artifacts.
We further extended SVD-based feature aggregation by introducing \textbf{SSVA}, which controls the integration of singular directions depending on the input, and \textbf{ID-RSVD}, which adjusts subspace estimation via input-dependent sketch generation.
Experiments on five benchmark datasets confirmed that the proposed method achieves higher classification accuracy than the baseline ViT.
These results suggest that incorporating SVD into feature maps is an effective approach to improving token aggregation in ViT.